\documentclass[11pt]{article}

\usepackage[margin=1in]{geometry}
\usepackage{amsmath, amssymb, amsthm}
\usepackage{booktabs}
\usepackage{graphicx}
\usepackage{natbib}
\usepackage{tikz}
\usepackage{pgfplots}
\pgfplotsset{compat=1.17}
\usetikzlibrary{arrows.meta, positioning, shapes.geometric, decorations.pathreplacing, calc}
\usepackage{xcolor}
\usepackage[protrusion=true,expansion=false]{microtype}
\usepackage{hyperref}
\hypersetup{colorlinks=true, linkcolor=blue!50!black, citecolor=blue!50!black, urlcolor=blue!50!black}
\usepackage{caption}
\usepackage{array}

\newtheorem{definition}{Definition}
\newtheorem{proposition}{Proposition}

\newtheorem{remark}{Remark}

\title{\vspace{-1em}\Large Learning to Compress Time-to-Control:\\[0.3em]
A Reinforcement Learning Framework for Chronic Disease Management}

\author{\normalsize Prabhjot Singh\thanks{Corresponding author: \texttt{prabhjot@joinaltitude.com}}, Abhishek Gupta, Chris Betz, Abe Flansburg,\\
Brett Ives, Sudeep Lama, Jung Hoon Son\\[0.6em]
\normalsize \textbf{Altitude}}
\date{\small Version: May 2026}

\begin{document}

\maketitle

\begin{abstract}
\noindent Reinforcement learning in healthcare has had mixed results, with reward sparsity, unreliable off-policy evaluation, and deployment-simulation gap as recurring failure modes. We argue that chronic disease management is structurally a more tractable RL setting than the acute-care problems the field has primarily studied, but only if the problem is formalized to exploit chronic care's properties. We propose such a formalization. The agent's objective is to compress time-to-control (TTC) under a tiered reward calibrated to the CMS ACCESS Model. Two quantities from our companion preference-learning paper~\citep{altitudep1} enter as load-bearing structural elements: the execution intensity $\varepsilon$ bounds action availability under a constrained MDP, and the clinician capability $\kappa$ weights offline-data transitions during RL training. Together they couple preference learning and RL into a two-loop architecture. We present simulation results on synthetic state machines for hypertension and type 2 diabetes. Capability-weighted offline RL outperforms uniform-weighted offline RL and the behavior policy by 15 percentage points on T2D TTC; the uniform-weighted formulation---the standard in existing healthcare RL---underperforms even the heterogeneous behavior policy. $\varepsilon$-aware policies generalize across deployment regimes while $\varepsilon$-naive policies do not.

\end{abstract}

\section{Introduction}
\label{sec:intro}

The application of reinforcement learning to clinical decision-making has produced an instructive body of work over the past decade. The Komorowski et al.\ AI~Clinician for sepsis~\citep{komorowski2018} represents the most-cited attempt and the one whose subsequent critique~\citep{gottesman2019, jeter2019} has shaped how the field thinks about RL in healthcare. Subsequent work on mechanical ventilation weaning~\citep{prasad2017}, dynamic treatment recommendation~\citep{wang2018}, and dosing optimization has refined the methodology, and recent surveys~\citep{yu2021, liu2020} have catalogued both the promise and the recurring difficulties.

The recurring difficulties are by now well-characterized. Reward signals in healthcare RL are typically sparse (mortality at hospital discharge), surrogate (time to extubation, length of stay), or imitative (matching the policy of physicians whose own performance is suboptimal). State spaces are high-dimensional and partially observed. Actions are confounded with severity-of-illness. Off-policy evaluation is unreliable when the behavior policy is poorly characterized. Most fundamentally, learned policies that look optimal in simulation often fail in deployment because the simulation does not capture the constraints under which the policy must actually execute---which clinicians will adopt the recommendation, which operational pathways will deliver the action, which patients will engage with the resulting plan.

Chronic disease management has not been the principal target of healthcare RL research, which we view as a missed opportunity. Chronic care has structural properties that make it a better fit for RL than the acute-care settings the field has focused on, provided three conditions are met: the reward signal is appropriately tiered, the action space is correctly typed, and execution intensity is treated as a learned constraint rather than an exogenous nuisance.

\subsection{Why chronic care fits RL}

Three structural properties favor RL in chronic disease management. \textbf{Decision density}: a patient with a chronic cardiometabolic condition under longitudinal management generates dozens to hundreds of decision points per year, orders of magnitude more than the per-patient decision count in a typical sepsis hospitalization. \textbf{Outcome observability}: chronic disease control is defined by quantitative biomarkers (BP, HbA1c, eGFR, LDL-C, weight) measured on a regular schedule, producing dense and informative reward signals rather than sparse binary mortality or readmission events. \textbf{Contractual alignment}: the CMS ACCESS Model~\citep{cmsaccess2026}, launching July~5, 2026, operationalizes outcome-aligned payment for chronic disease management with explicit Control and Minimum Improvement targets per condition---a payer-validated reward structure in production, conferring external validity that healthcare RL has rarely had. The combination is favorable for RL; the reason chronic care has not been a more prominent RL target is that the field has imported reward structures and constraint assumptions from acute-care settings that do not exploit it.

\subsection{Contributions and positioning}

The contribution of this paper is at the level of \emph{problem formulation}, not algorithm: we specify the chronic care RL problem so that existing offline RL methods, applied to it, have a chance of producing deployable policies.

First (Section~\ref{sec:related}), we review the existing healthcare RL literature and identify the structural reasons it has produced limited deployable success. The failure modes the Komorowski sepsis work and its critique~\citep{komorowski2018, gottesman2019, jeter2019} surfaced---reward gaming under behavior-cloning constraints, off-policy evaluation unreliability, deployment-simulation gap---are the same failure modes a chronic care RL formulation must avoid by design.

Second (Section~\ref{sec:substrate}), we formalize the substrate: patient state machines with typed actions. The action space partitions into a clinical layer (medication intensification, lab orders, referrals) and an operational layer (scheduling, outreach, follow-through). The typing matters because the two layers have different causal structure; treating them as a single undifferentiated action space, as most healthcare RL work has, obscures the variance that operational execution contributes to outcomes.

Third (Section~\ref{sec:reward}), we formalize the TTG/TTO/TTC reward hierarchy as first-passage times on the patient state machine. The tiered reward structure addresses the sparsity problem of single-endpoint formulations and is calibrated to the CMS ACCESS Model~\citep{cmsaccess2026}, which operationalizes an analogous tier (Minimum Improvement / Control) in production. The tier weights inherit external validity from a payer-published, contractually-binding structure.

Fourth (Section~\ref{sec:constraint}), and most distinctively, we couple the framework to clinician preference learning. Two quantities derived in our companion paper~\citep{altitudep1}---the execution intensity $\varepsilon = \varepsilon_{\mathrm{cap}} \cdot \varepsilon_{\mathrm{will}}$ and the clinician capability $\kappa$---enter this paper as load-bearing structural elements. The execution intensity parameterizes the action-availability constraint of a constrained MDP. The capability score parameterizes a transition-weighting term in the offline RL training objective. Together they define a two-loop architecture: an outer loop in which preference learning continuously updates posteriors over $\varepsilon$ and $\kappa$ from accumulated clinician override data, and an inner loop in which the RL agent uses these as inputs to its training objective. This coupling is the framework's central technical claim.

Fifth (Section~\ref{sec:harness}), we describe a supervisory layer---an engagement harness---that bounds agent autonomy. The harness routes high-uncertainty or high-risk decisions to clinician review while permitting autonomous execution where it is safe. We describe the harness abstractly, identify its required functional components, and discuss a concrete design we have proposed for the post-discharge engagement domain as one instantiation of the abstract specification.

Section~\ref{sec:empirical} presents empirical anchoring: actuarial modeling that supports the framework's economic implications, and a simulation study that tests the two-loop architecture directly. The simulation establishes that capability-weighted offline RL outperforms uniform-weighted offline RL and the behavior policy baseline; we treat this as confirmation that the P1 coupling is the load-bearing innovation the framework claims.

\section{Related work and the failure modes to avoid}
\label{sec:related}

Healthcare RL has progressed substantially since the early sepsis work, but several failure modes recur and are worth naming explicitly because the chronic care formulation must avoid them by design.

\textbf{Reward gaming under behavior-cloning constraints.} The Komorowski AI~Clinician~\citep{komorowski2018} learned a policy on retrospective ICU data using a discrete state-action representation and reward derived from 90-day mortality. Subsequent analysis~\citep{jeter2019} demonstrated that the learned policy was sensitive to the specific discretization of state space and that its apparent superiority over clinician policy was partly an artifact of the off-policy evaluation procedure. The deeper issue, articulated in the guidelines that followed~\citep{gottesman2019}, is that any RL agent trained against a sparse end-state reward will exploit whatever shortcuts the data permit, including shortcuts that reflect data-generation artifacts rather than clinical mechanism. This is not specific to sepsis; it is the standard reward-hacking problem in offline RL. Dense, intermediate, mechanism-anchored rewards are the principal defense.

\textbf{Off-policy evaluation unreliability.} Healthcare RL is almost always offline RL. The behavior policy that generated the data is a mixture of clinician decisions, organizational defaults, and patient-level idiosyncrasies, and is not directly characterized. Importance-sampling-based off-policy evaluation~\citep{levine2020} has high variance under these conditions, and conservative offline RL methods~\citep{kumar2020, fujimoto2019} address some but not all of the reliability problem. The chronic care formulation we develop is more amenable to off-policy evaluation than the sepsis case because the decision points are denser, the outcomes are more frequently observed, and the behavior policy is partially characterized through the clinician override data central to our companion paper~\citep{altitudep1}---but the difficulty remains real and we treat it explicitly in Section~\ref{sec:empirical}.

\textbf{Deployment-simulation gap.} A learned policy that performs well on retrospective data may fail in deployment because the deployment environment differs from the data-generating process in ways the simulation did not capture. For chronic care, the principal source of this gap is execution intensity: a policy may recommend a medication intensification that the operational chain cannot deliver, or an alert the clinician will not act on. The simulation, trained on data where these constraints were implicit, may not surface the failure. We make execution intensity an explicit constraint on action availability (Section~\ref{sec:constraint}) precisely to close this gap.

\textbf{State space contamination.} High-dimensional state representations in healthcare RL frequently include features that are themselves consequences of the action being evaluated, producing a form of leakage that biases evaluation toward the behavior policy. The patient state machine formulation (Section~\ref{sec:substrate}) addresses this by defining state in terms of clinically interpretable, condition-specific variables that are causally upstream of the action choice rather than downstream of it.

\section{Patient state machines and typed actions}
\label{sec:substrate}

\subsection{State machines per condition}

For each chronic condition $c$ and patient $i$, we model the patient's clinical state as evolving on a continuous-time process $S_{i,c}(t) \in \mathcal{S}_c$, where $\mathcal{S}_c$ is the condition-specific state space. The state space partitions into a controlled region $\mathcal{C}_c$ and an uncontrolled region $\mathcal{U}_c = \mathcal{S}_c \setminus \mathcal{C}_c$, defined by clinical thresholds that derive from current treatment guidelines:

\begin{itemize}
\setlength{\itemsep}{0.2em}
\item \textbf{Hypertension}: $\mathcal{C}_{\mathrm{HTN}} = \{(\mathrm{SBP}, \mathrm{DBP}) : \mathrm{SBP} < 130, \mathrm{DBP} < 80\}$ on guideline-directed antihypertensive therapy~\citep{whelton2018}, with thresholds adjusted downward in higher-risk profiles (CKD, T2D, ASCVD).
\item \textbf{Type 2 diabetes}: $\mathcal{C}_{\mathrm{T2D}} = \{\mathrm{HbA1c} : \mathrm{HbA1c} < 7.0\%\}$ on therapy, with adjustment for frailty, hypoglycemia risk, and life expectancy.
\item \textbf{Dyslipidemia / ASCVD risk}: $\mathcal{C}_{\mathrm{LDL}} = \{\mathrm{LDL\text{-}C} : \mathrm{LDL\text{-}C} < 70\,\mathrm{mg/dL}\}$ on guideline-indicated statin therapy at appropriate intensity, with adjunct lipid-lowering therapy as needed; threshold loosens in lower-risk primary-prevention profiles. Lipid lowering is the cardio-kidney-metabolic lever with the most direct ASCVD-event reduction in CKD and T2D populations.
\item \textbf{HFrEF}: no clean numerical control region. The relevant target state is being on full guideline-directed medical therapy at target dose~\citep{heidenreich2022}; ``control'' is replaced by absorption into a maintenance state.
\item \textbf{CKD}: no clean control region. Target state is therapeutic regimen completion (RAS blockade, SGLT2 inhibitor where eligible, finerenone where indicated)~\citep{kdigo2024} with biomarker stabilization.
\end{itemize}

The state machine is observed only at measurement times $\tau_{i,c}^{(1)} < \tau_{i,c}^{(2)} < \cdots$ generated by the encounter and laboratory schedule. At each measurement time, the agent observes $S_{i,c}(\tau_{i,c}^{(k)})$ and may choose an action.

\subsection{The two-layer action taxonomy}

Actions in chronic care decision-making partition into two layers with distinct causal structure.

\begin{definition}[Two-layer action taxonomy]
\label{def:actions}
The action space $\mathcal{A}$ partitions as $\mathcal{A} = \mathcal{A}_{\mathrm{clin}} \cup \mathcal{A}_{\mathrm{ops}}$, where:
\begin{itemize}
\item $\mathcal{A}_{\mathrm{clin}}$ is the \emph{clinical layer}: actions that target the patient's clinical state directly. Medication initiation and titration, laboratory orders, specialist referrals, intensification or de-intensification decisions, diagnostic workup.
\item $\mathcal{A}_{\mathrm{ops}}$ is the \emph{operational layer}: actions that target whether and when clinical actions execute. Scheduling, appointment confirmation, patient outreach, follow-through tracking, escalation when an expected action does not occur.
\end{itemize}
\end{definition}

The two layers have different relationships to outcomes. Clinical actions $a \in \mathcal{A}_{\mathrm{clin}}$ act on the patient's biological process: a recommendation to add an SGLT2 inhibitor, if executed, changes the trajectory of $S_{i,c}$ through pharmacological mechanism. Operational actions $a \in \mathcal{A}_{\mathrm{ops}}$ act on whether a clinical action gets executed: a confirmed visit, a refilled prescription, a successful outreach call all increase the probability that the next clinical action in the sequence is taken. Operational actions are causal mediators of outcomes, but they mediate through the clinical layer rather than directly.

The standard healthcare RL formulation collapses these into a single action space, which makes two important quantities unidentifiable. First, the contribution of operational execution to outcomes cannot be separated from clinical decision quality, even though the two have very different causal mechanisms and are determined by different deployment infrastructure. Second, the constraint structure described in Section~\ref{sec:constraint}---which differs across the two layers because operational capacity and clinician willingness affect them differently---cannot be expressed. We treat the two layers as distinct throughout the formulation.

\subsection{Observation density and the measurement process}

The measurement schedule $\{\tau_{i,c}^{(k)}\}$ is itself a part of the state-action environment that the agent operates in. Observation density---how frequently $S_{i,c}$ is sampled---governs the rate at which state transitions become detectable, and therefore the rate at which the agent can update its action policy. For conditions with home-monitoring infrastructure (blood pressure cuffs, continuous glucose monitors, smart scales), observation density can be daily or near-continuous. For conditions reliant on clinical encounters (HbA1c at quarterly intervals, eGFR at variable cadence), observation density is sparse.

\begin{remark}
Observation density is partially under the agent's control. Operational actions that schedule lab orders or initiate home-monitoring increase $\{\tau_{i,c}^{(k)}\}$ density and therefore improve the signal available for subsequent decisions. This creates a meta-level decision: the agent should sometimes prefer operational actions that increase observability over clinical actions whose effect cannot yet be observed. This is the chronic care analog of the explore-exploit tradeoff and is a central design consideration we return to in Section~\ref{sec:constraint}.
\end{remark}

\subsection{Stall states}
\label{sec:stalls}

A patient state machine generates two qualitatively distinct kinds of trajectories. In a \emph{progressing} trajectory, the patient moves through the milestone sequence (TTG, TTO, TTC) at a pace consistent with the underlying biological response curves to titrated therapy. In a \emph{stalled} trajectory, the patient remains in $\mathcal{U}_c$ or in an intermediate state without progressing for substantially longer than the response curves would predict. Stalls are the trajectory shapes that drive the long right tail of TTC distributions in deployed chronic care platforms; they are also the trajectory shapes most amenable to intervention, because a stall by definition indicates that whatever the current action policy is doing is not producing the expected progression.

We define three stall types as derived first-passage timeouts on the state machine:

\begin{definition}[Stall states]
\label{def:stall}
For patient $i$ in condition $c$ with index uncontrolled time $T_0^{i,c}$, and using the milestone times of Definition~\ref{def:hierarchy}, define three stall indicators as functions of the running time $t$ since $T_0^{i,c}$:
\begin{itemize}
\item $\mathrm{Stall}_{\mathrm{G}}(t) = \mathbb{1}[t > \tau_G \wedge \mathrm{TTG}_{i,c} > t]$ (\emph{progress stall}): the patient has remained in $\mathcal{U}_c$ for time exceeding $\tau_G$ without yet hitting the TTG milestone.
\item $\mathrm{Stall}_{\mathrm{O}}(t) = \mathbb{1}[t > \mathrm{TTG}_{i,c} + \tau_O \wedge \mathrm{TTO}_{i,c} > t]$ (\emph{intermediate stall}): the patient has hit TTG but has remained in the partially-controlled region for time exceeding $\tau_O$ since TTG without progressing to TTO.
\item $\mathrm{Stall}_{\mathrm{R}}(t) = \mathbb{1}[\mathrm{TTC}_{i,c} \leq t \wedge S_{i,c}(t) \in \mathcal{U}_c \wedge \text{loss persisted} \geq \tau_R]$ (\emph{regression stall}): the patient previously achieved TTC but has fallen back into $\mathcal{U}_c$ for time exceeding the recovery window $\tau_R$, after which loss-of-control is treated as a stall rather than transient noise.
\end{itemize}
The TTG and TTO times here are durations from $T_0^{i,c}$ as defined in Definition~\ref{def:hierarchy}, with the convention that $\mathrm{TTG}_{i,c} > t$ means TTG has not yet been observed by elapsed time $t$.
\end{definition}

The thresholds $\tau_G, \tau_O, \tau_R$ are calibrated to the per-condition response curves: a hypertension patient on guideline-directed dual therapy should achieve TTG within 6--8 weeks, so $\tau_G = 8$ weeks for HTN is a reasonable starting calibration. For T2D under metformin titration, $\tau_G = 16$ weeks reflects the slower glycemic response. These thresholds are not new clinical knowledge; they are the well-established response-time expectations against which therapeutic inertia is itself defined~\citep{phillips2001, okonofua2006, khunti2018}.

Stall states are central to the framework's prescriptive content. Three properties make them particularly important. First, they are \emph{detectable} as a derived feature of the state trajectory: the agent does not need richer state representation to recognize a stall, only the running time-since-index counter and the milestone history. Second, they are \emph{differentially actionable}: a stall is exactly the state in which the existing clinical action stream has failed to produce expected progression, which means a status-quo policy is by definition the wrong policy. Third, they are \emph{the natural trigger for autonomous operational intensification}: when a patient stalls, the framework's prescription is for the agent to autonomously increase operational action density (outreach, voice agent contact, scheduling, observability-raising actions) and concurrently to escalate the clinical decision through the harness with elevated priority.

This directive interacts with each subsequent section of the framework. In Section~\ref{sec:reward}, stall persistence enters the reward structure as a candidate negative term: time spent in $\mathrm{Stall}_{\mathrm{G}}$ accumulates a small per-week penalty that gives the agent direct signal to disprefer policies leading to stalls. In Section~\ref{sec:constraint}, stall states justify autonomous operational action even at low $\varepsilon$, because the cost of continued stall is itself accumulating in the reward signal; the action-availability constraint relaxes when the alternative is a confirmed stall. In Section~\ref{sec:harness}, stall duration enters the routing function as a third trigger alongside uncertainty and risk: a long stall escalates to clinician review with elevated priority, while autonomous operational responses (outreach, scheduling) execute without escalation since they fall within the harness's autonomous-operational envelope. The empirical demonstration of stall-aware policies---specifically, whether they compress TTC more than stall-naive policies in patient subpopulations where therapeutic inertia is most pronounced---is a deployment-side prediction we identify as a target for prospective evaluation.

\subsection{Cross-condition prioritization in the joint patient state}
\label{sec:prioritization}

The condition-specific framing above is a notational convenience; in deployment, a patient is in a \emph{joint} state across the conditions they carry, $S_i(t) = (S_{i,c_1}(t), \ldots, S_{i,c_K}(t))$, with stall indicators defined per-condition but a single global trajectory of disease risk. The agent's prescriptive question is therefore not only ``compress TTC for condition $c$'' but ``which stalled condition state, addressed now, most reduces this patient's global cardio-kidney-metabolic risk over the next decision horizon.''  In a patient with comorbid HTN, T2D, and CKD, an SBP stall and an LDL-C stall may carry different marginal-risk weights depending on baseline ASCVD risk, eGFR trajectory, and prior event history; the operational layer's finite capacity at any decision point is allocated more effectively when this prioritization is explicit. We therefore treat condition-state prioritization as a first-class output of the agent: at each decision point, the agent ranks the patient's currently stalled or near-stalled conditions by predicted marginal global-risk reduction per available action, and routes both clinical and operational action capacity to the top-ranked condition first.

A consequence of this framing is that some clinical actions have value precisely because they act across condition states simultaneously. SGLT2 inhibitors are the clearest current example: a single pharmacologic action contributes to TTC compression in HFrEF, CKD, and T2D in parallel, and does so at different points in each condition's state machine~\citep{heidenreich2022, kdigo2024}. Under cross-condition prioritization, such actions receive substantially elevated value because they discharge multiple stalled trajectories at once for the cost of a single execution event, and we expect them to dominate prioritization output in the comorbid-patient regime. Statins occupy a similar position for the dyslipidemia / ASCVD axis. The framework's preference for cross-condition actions is not hand-coded; it falls out of the joint-state reward structure when the per-condition tiered rewards of Section~\ref{sec:reward} accumulate concurrently for a single execution. Operationalizing this in production requires the patient state encoder $\phi(s)$ (Section~\ref{sec:discussion}) to represent the joint state faithfully and the reward hierarchy to credit concurrent milestone progression correctly; both are deployment problems the framework constrains but does not by itself solve.

\section{The TTG/TTO/TTC reward hierarchy}
\label{sec:reward}

\subsection{Three first-passage times}

We define three reward-relevant time intervals, each measured from an index uncontrolled observation.

\begin{definition}[Time-to-control hierarchy]
\label{def:hierarchy}
Let $T_0^{i,c}$ denote the index time at which patient $i$ is first observed in $\mathcal{U}_c$ within the platform observation window. Define three first-passage times (random variables that may take the value $\infty$ if the milestone is never reached within the observation window):
\begin{itemize}
\item $\mathrm{TTG}_{i,c}$: time from $T_0^{i,c}$ to the first observation of a \emph{clinically meaningful step-change} in the direction of $\mathcal{C}_c$, where ``meaningful'' is defined per condition by a biomarker reduction threshold (e.g., $\geq 15$~mmHg SBP reduction for HTN, $\geq 1.0\%$ HbA1c reduction for T2D).
\item $\mathrm{TTO}_{i,c}$: time from $T_0^{i,c}$ to the first observation of \emph{either} (a)~a substantial biomarker change short of full control (e.g., stage transition for HTN, $\geq 30\%$ reduction in albuminuria for CKD), \emph{or} (b)~a structural therapeutic milestone short of control (e.g., full titration to target dose, completion of an indicated medication class).
\item $\mathrm{TTC}_{i,c}$: time from $T_0^{i,c}$ to the first observation of $S_{i,c} \in \mathcal{C}_c$ that is sustained across a confirmation window $\Delta$.
\end{itemize}
\end{definition}

The three intervals are nested where finite: $\mathrm{TTG}_{i,c} \leq \mathrm{TTO}_{i,c} \leq \mathrm{TTC}_{i,c}$ on the event that all three are observed within the platform window. For conditions without a clean control region (HFrEF, CKD), $\mathrm{TTC}$ is replaced by the time to absorption into a guideline-directed maintenance state.

\subsection{Tiered reward and ACCESS anchoring}

The reward function combines the three intervals into a tiered shaping structure. At each measurement time $\tau_{i,c}^{(k)}$, the agent receives reward
\begin{multline}
R(s_{i,c}^{(k)}, a_{i,c}^{(k)}, s_{i,c}^{(k+1)}) = w_G \cdot \mathbb{1}[\tau_{i,c}^{(k+1)} = T_0^{i,c} + \mathrm{TTG}_{i,c}] + w_O \cdot \mathbb{1}[\tau_{i,c}^{(k+1)} = T_0^{i,c} + \mathrm{TTO}_{i,c}] \\
+ w_C \cdot \mathbb{1}[\tau_{i,c}^{(k+1)} = T_0^{i,c} + \mathrm{TTC}_{i,c}] - \rho(a_{i,c}^{(k)}),
\label{eq:reward}
\end{multline}
where each indicator fires exactly once per trajectory at the time of first passage to the corresponding milestone (and zero on all other observations), $w_G \leq w_O \leq w_C$ are tier weights, and $\rho(a)$ is a per-action cost capturing clinician burden, patient burden, and operational cost. Throughout this paper, $\rho$ denotes per-action cost; the symbol $\kappa$ is reserved for clinician capability (Section~\ref{sec:kappa}).

The tier weights are not analyst-chosen. The CMS ACCESS Model~\citep{cmsaccess2026} operationalizes a two-tier version of this hierarchy in production: each ACCESS performance measure has a paired Control target and Minimum Improvement target, and outcome-aligned payment is earned when at least 50\% of aligned beneficiaries achieve either. The Minimum Improvement targets are precisely TTG-tier definitions---``a minimum change that is clinically meaningful, not the beneficiary's ultimate clinical goal''---specified as $\geq 15$~mmHg SBP reduction, $\geq 1.0\%$ HbA1c reduction (CKM track), $\geq 30$~mg/dL LDL-C reduction, or $\geq 5\%$ weight reduction. Because ACCESS allocates roughly equal contractual weight to Minimum Improvement and Control attainment within a 12-month care period, $w_G \approx w_C$ with $w_O$ at or above this level is a defensible starting calibration; this is consistent with the weak ordering above and motivated by contractual alignment rather than by reward-shaping theory. A learned policy that compresses TTG and TTC under this reward structure is by construction aligned with the contractual incentives of any ACCESS-participating organization.

\begin{proposition}[Density of the tiered reward signal]
\label{prop:density}
Under the tiered reward of Equation~(\ref{eq:reward}), each patient trajectory in which all three milestones are eventually observed contributes exactly three non-zero positive reward events (one each for TTG, TTO, TTC) plus a non-positive sequence of action costs. Trajectories on which only TTG is hit contribute one positive reward event; trajectories on which no milestone is hit contribute zero. Across a heterogeneous population at typical chronic care platform performance ($\sim 50\%$ TTG and $\sim 20\%$ TTC achievement within twelve months), the expected number of non-zero positive reward events per patient-year is $\sim 0.7$--$1.5$. Augmenting the reward with intermediate state-change events (control loss, partial-response transitions, or the stall-state events of Section~\ref{sec:stalls}) raises this density to $\sim 5$--$30$ events per patient-year for HTN under home monitoring. By contrast, single-endpoint TTC-only reward yields $\sim 0.2$ events per patient-year, and single-endpoint mortality reward yields $\sim 10^{-2}$ events per patient-year.
\end{proposition}

The order-of-magnitude differences in reward density are the principal mechanism by which the tiered reward improves credit assignment. Against a single-endpoint TTC-only reward (the natural chronic-care comparison), the tiered reward of Equation~(\ref{eq:reward}) provides on the order of $5\times$--$10\times$ more positive reward events per patient-year, with an additional order of magnitude available if intermediate state-change events are credited. Against single-endpoint mortality reward (the chronic-care analog of the standard sepsis RL formulation), the tiered reward provides on the order of $10^2$--$10^3 \times$ more learning signal per patient-year. Most reward functions in the existing healthcare RL literature are analyst proposals validated only by clinical face validity; the TTG/TTO/TTC reward inherits its validity from CMS itself, which is the unusual structural property that makes chronic care RL more tractable than acute-care RL formulations.

\section{Coupling to clinician preference learning}
\label{sec:constraint}

The framework's most distinctive feature is its tight coupling to the preference-learning machinery developed in our companion paper~\citep{altitudep1}. Two quantities from that paper enter this paper as load-bearing structural elements: the execution intensity $\varepsilon$ as an action-availability constraint, and the clinician capability $\kappa$ as a transition-weighting signal during offline RL training. Together they define a two-loop architecture in which clinician outcome data continuously informs both what the agent can do (via $\varepsilon$) and from whom the agent learns (via $\kappa$).

\subsection{Execution intensity as an action-availability constraint}
\label{sec:eps}

A standard RL formulation assumes the agent has unconstrained access to its action space: at each decision point, any action $a \in \mathcal{A}$ is available, and the only question is which to choose. This assumption fails in healthcare. A platform that recommends an SGLT2 inhibitor cannot itself execute the recommendation; the recommendation must be acted upon by a clinician, and the action must be operationally completed (prescription generated, sent, filled, taken). At each link in the chain, the recommendation may not execute. The probability that it does execute is the execution intensity $\varepsilon$, which~\citet{altitudep1} decomposes as
\begin{equation}
\varepsilon = \varepsilon_{\mathrm{cap}} \cdot \varepsilon_{\mathrm{will}}
\label{eq:eps}
\end{equation}
where $\varepsilon_{\mathrm{cap}}$ is the operational capacity (probability the operational pathway successfully delivers the action conditional on a clinician decision to act) and $\varepsilon_{\mathrm{will}}$ is the clinician willingness (probability of clinician initiation conditional on operational availability). The product form assumes conditional independence of capacity and willingness given the agent's recommendation; \citet{altitudep1} discusses this assumption and the conditions under which it holds approximately.

The agent's environment is most naturally formalized as a constrained Markov decision process~\citep{altman1999} in which $\varepsilon$ is a (partially observable) state variable that determines action availability. Concretely, we treat $\varepsilon$ as a component of the augmented state $s = (s_{\mathrm{clin}}, \varepsilon)$ where $s_{\mathrm{clin}}$ is the patient's clinical state and $\varepsilon$ is the agent's posterior estimate of execution intensity at the deployment context. Different actions can have different execution profiles---a voice-agent contact has a higher and more reliable $\varepsilon$ than a same-day specialist referral---so we let $\varepsilon$ depend on the action under consideration. The available action subset at state $s$ is
\begin{equation}
\mathcal{A}^{\mathrm{avail}}(s) = \{a \in \mathcal{A} : \hat{\varepsilon}(s, a) \geq \varepsilon_{\min}(a)\}
\label{eq:availability}
\end{equation}
where $\hat{\varepsilon}(s, a)$ is the agent's estimate of execution intensity for action $a$ at state $s$ (drawn from the outer-loop preference learning of Section~\ref{sec:kappa}), and $\varepsilon_{\min}: \mathcal{A} \to [0, 1]$ is an action-specific availability threshold calibrated from the response-window requirements of each action (a recommendation that has $<5\%$ chance of being completed within the action's clinically meaningful window has effectively zero availability). This is not standard reward-constrained RL where a constraint cost is subtracted from reward; it is action-availability-constrained RL where the agent's effective action space is itself a function of the partially-observable execution-intensity state. The agent must reason over which actions are likely to execute given the current state, not merely which actions would be optimal if execution were free.

A standard offline RL pipeline trained without explicit treatment of $\varepsilon$ will learn a policy that implicitly bakes in the average execution intensity of the training data. When deployed in a setting with different $\varepsilon$, the policy will systematically over- or under-recommend---not because it is wrong on the patients it sees, but because the patients it sees are not the patients the policy was implicitly designed for. The constrained MDP formulation addresses this by making $\varepsilon$ explicit in the agent's decision process.

The state-dependence of $\varepsilon$ is also patient-specific, not just clinician- and deployment-specific. Patient social and structural factors---food insecurity, housing instability, transportation barriers, financial strain---affect both $\varepsilon_{\mathrm{cap}}$ (whether the operational pathway can deliver the action: a referral the patient cannot get to has effectively zero capacity) and $\varepsilon_{\mathrm{will}}$ (clinicians may rationally adjust escalation aggressiveness to predicted patient follow-through, and unconscious adjustments in the same direction are documented in the disparities literature). A framework that treats $\varepsilon$ as patient-state-dependent must therefore include patient social and structural state in $\hat{\varepsilon}(s, a)$. This raises an equity question that the framework cannot resolve internally: should the agent learn from data in which clinicians have made differentially aggressive recommendations to patients with different SDOH profiles, or should the policy explicitly correct for that differential? We treat this as a deployment-policy question that requires explicit organizational stance rather than as a structural feature of the framework, but flag that any production deployment must address it.

Execution intensity is not exogenous to the agent. The operational layer $\mathcal{A}_{\mathrm{ops}}$ contains actions that directly raise $\varepsilon_{\mathrm{cap}}$, and $\varepsilon_{\mathrm{will}}$ is shaped by the clinician's accumulated experience with the agent's prior recommendations. As $\varepsilon$ rises through learning, the agent's effective action space expands and the policies it can pursue widen. This compounding effect---an RL agent in chronic care exhibiting accelerating performance over its operational lifetime as $\varepsilon$ rises---is a property that static analyses of policy quality at fixed $\varepsilon$ do not capture.

\subsection{Capability weighting as a transition-weighting signal}
\label{sec:kappa}

A standard offline RL pipeline also assumes that all transitions in the training data are equally informative. This assumption fails when the data is generated by clinicians of widely varying capability. A policy trained on uniform-weighted transitions imitates the population mean of clinician behavior; in chronic care, the population mean reflects significant therapeutic inertia~\citep{phillips2001, okonofua2006, khunti2018} and is a poor target. The agent that imitates the mean inherits the mean's failure modes.

\citet{altitudep1} formalizes a Bradley-Terry preference model over clinicians in which a capability score $\kappa_i = \kappa_i^{\mathrm{exec}} + \kappa_i^{\mathrm{align}}$ is inferred per clinician from outcome data. The execution component $\kappa_i^{\mathrm{exec}}$ captures the clinician's effectiveness at clinical decision-making; the alignment component $\kappa_i^{\mathrm{align}}$ captures their consistency with VBC objectives. Together they parameterize a posterior distribution over clinician capability that is updated as outcome evidence accumulates.

We use $\kappa$ as a transition-weighting signal during offline RL training. For each transition $(s, a, r, s')$ generated by clinician $i$, the gradient contribution is scaled by
\begin{equation}
w_i = \exp(\beta \kappa_i)
\label{eq:weighting}
\end{equation}
where $\beta \geq 0$ is a temperature parameter. At $\beta = 0$, all transitions are equally weighted (standard offline RL). At $\beta \to \infty$, the policy reduces to imitation of the highest-capability clinician. Intermediate $\beta$ produces a soft preference for high-capability transitions while retaining coverage from the broader population.

The mechanism is neither novel as a generic technique (weighted offline RL is well-studied) nor specific to chronic care in its mathematical form. The contribution is its anchoring: capability is inferred from clinician override data using the framework of~\citet{altitudep1}, which addresses the standard offline RL problem of unidentified behavior policy by explicitly modeling clinician-level variation. The Bradley-Terry preference structure provides a principled way to estimate $\kappa$ from data that real chronic care platforms already generate.

\subsection{The two-loop architecture}
\label{sec:twoloop}

Together, $\varepsilon$ and $\kappa$ define a two-loop architecture in which preference learning and reinforcement learning are tightly coupled.

\textbf{Outer loop (preference learning).} As clinician override data accumulates, the framework of~\citet{altitudep1} continuously updates the posterior over $(\varepsilon_{\mathrm{cap}}, \varepsilon_{\mathrm{will}}, \kappa_i)$ at the patient, clinician, and platform levels. This loop operates over the timescale at which override data accumulates---months in current deployments---and produces calibrated estimates of execution intensity and clinician capability.

\textbf{Inner loop (RL training).} The RL agent uses these estimates as direct inputs to its training objective: $\hat{\varepsilon}$ enters the action-availability constraint of Equation~(\ref{eq:availability}); $\kappa$ enters the transition weighting of Equation~(\ref{eq:weighting}). Let $\mathcal{D} = \{(s_j, a_j, r_j, s'_j, i_j)\}_{j=1}^N$ denote the offline dataset of transitions tagged with the clinician identity $i_j$ that generated each. The agent's training objective is
\begin{equation}
\mathcal{L}(\theta) = \mathbb{E}_{(s, a, r, s', i) \sim \mathcal{D}} \left[ w_i \cdot \big( r + \gamma \max_{a' \in \mathcal{A}^{\mathrm{avail}}(s')} Q_{\bar{\theta}}(s', a') - Q_\theta(s, a) \big)^2 \right]
\label{eq:objective}
\end{equation}
where $w_i = \exp(\beta \kappa_i)$ is the capability weight from Equation~(\ref{eq:weighting}), the inner max is restricted to executable actions per the action-availability function, and $Q_{\bar{\theta}}$ denotes a target network (or the previous-iteration tabular estimate in the discrete case) used to stabilize the Bellman target~\citep{mnih2015}. In practice the weighting can be implemented either by scaling the per-transition gradient or by importance sampling transitions proportionally to $w_i$; the two are equivalent up to variance and we use importance sampling in our simulation. This loop operates over the timescale of agent retraining---weeks to months---and produces an updated policy parameterized by $\theta$.

The coupling is bidirectional. The outer loop's estimates parameterize the inner loop's training. The inner loop's policy, when deployed, generates new override data that updates the outer loop's estimates. This is the operational realization of the framework's central claim: that an RL agent for chronic care must be coupled to a preference-learning system that continuously identifies which clinicians and which deployment contexts produce the outcomes the agent should learn to reproduce.

The simulation in Section~\ref{sec:simulation} tests this two-loop architecture directly. We compare uniform-weighted offline RL ($\beta = 0$) against capability-weighted offline RL ($\beta > 0$) under matched data, matched algorithm, and matched reward structure. The capability-weighted variant should outperform the uniform-weighted one when the data is generated by clinicians of heterogeneous capability, which is the realistic case.

\section{The engagement harness as supervisory layer}
\label{sec:harness}

\subsection{Why a supervisory layer is necessary}

The framework developed so far---tiered reward, typed actions, execution-intensity-constrained MDP---defines what an agent should optimize but does not yet specify what bounds its autonomy in the high-uncertainty or high-risk decisions where autonomous action would be inappropriate. The need for such a supervisory layer is a standard concern in the broader AI safety literature~\citep{amodei2016, saunders2018} and is particularly acute in healthcare because the cost of a wrong autonomous action is measured in patient harm rather than only in lost reward. We refer to the supervisory layer in our framework as the \emph{engagement harness}: it is structured to preserve the operating envelope where the agent is genuinely useful while constraining it where it is not yet trusted.

\subsection{Functional components}

A supervisory harness for a chronic care RL agent must support four functional components.

\textbf{Uncertainty quantification.} At each decision point, the harness requires an estimate of the agent's confidence in its proposed action. This may be derived from posterior variance over a Bayesian RL model, ensemble disagreement across multiple trained policies, distance from the training distribution in a learned representation space, or an explicit confidence head trained alongside the policy. The specific method is implementation-dependent; what matters for the harness is that uncertainty is quantified and thresholded.

\textbf{Atypicality detection.} Decisions for patients whose state trajectory falls outside the training distribution---unusual comorbidity patterns, unusual response trajectories, demographic groups underrepresented in training---should be routed to clinician review independent of the agent's own confidence estimate. Atypicality detection is a separate function from uncertainty quantification: the agent may be confident in a recommendation that is wrong because it has never seen this patient type before. Out-of-distribution detection methods, fairness-aware monitoring, and subgroup performance tracking all serve this function.

\textbf{Risk-threshold triggers.} Some decisions cross a risk threshold above which clinician review is required regardless of agent confidence: high-dose medication initiation, decisions for patients with active acute issues, decisions that conflict with documented patient preferences. These thresholds are exogenously specified by clinical and regulatory constraints, not learned by the agent.

\textbf{Cadence and consent enforcement.} Operational layer actions ($\mathcal{A}_{\mathrm{ops}}$) are subject to hard runtime constraints: contact frequency limits (TCPA), time-of-day windows, patient consent state, escalation rules. These are non-negotiable rules enforced by the harness, not soft preferences the policy can trade off against expected reward. Runtime-enforced safety is the locus where AI systems and clinical operations actually meet, and a harness that delegates these constraints to the policy rather than enforcing them at the runtime level inherits the policy's failure modes as safety failures.

\subsection{Routing decisions}

At each decision point, the harness routes the agent's proposed action to one of three outcomes:

\begin{enumerate}
\setlength{\itemsep}{0.2em}
\item \textbf{Autonomous execution.} The action is in $\mathcal{A}_{\mathrm{ops}}$, falls below the risk threshold, has high agent confidence, the patient is in-distribution, and cadence/consent constraints are satisfied. The harness executes the action without clinician involvement. Examples: scheduling a follow-up visit, sending a refill reminder, initiating outreach for a missed appointment.
\item \textbf{Clinician review with default.} The action is in $\mathcal{A}_{\mathrm{clin}}$ and is at or near the risk threshold, or the agent's confidence is moderate. The harness surfaces the action to the clinician with a recommended decision and proceeds with the default if the clinician does not respond within a specified window. Examples: standard medication intensifications for guideline-supported indications.
\item \textbf{Clinician decision required.} The action crosses the risk threshold, the agent's confidence is low, the patient is out-of-distribution, or the action is in a category that requires human judgment by exogenous specification. The harness surfaces the decision to the clinician without a default; no action is taken until the clinician has reviewed. Examples: novel medication initiations, decisions for patients with unusual trajectories, decisions that conflict with patient-stated preferences.
\end{enumerate}

The routing function is the mechanism by which the agent's autonomy is graduated as trust accumulates. As the agent's track record extends, as patient-population coverage expands, as confidence calibration improves, more decisions move from category 3 to 2 to 1. This is the dynamic version of the autonomy claim: the harness does not statically determine the agent's operating envelope; it adjusts the envelope based on the agent's demonstrated reliability.

\subsection{A concrete instantiation: post-discharge engagement}

To make the abstract harness specification concrete, we describe a design we have proposed for post-discharge transitional care for high-risk patients. In this domain the operational layer's task is to reach a patient within the 48-hour window that determines whether transitional care milestones (medication reconciliation, follow-up visit completion, symptom check-in) are met. The state space is contact history, available channels, demographic and clinical characteristics, and current cadence/consent state. The action space is the choice of channel (AI voice agent, SMS, patient portal, scheduler bot, nurse callback, community health worker visit, family proxy outreach) and timing. The reward signal is engagement outcome within the contractually relevant window.

This domain is a useful instantiation for three reasons. The action space is discrete and well-bounded, simplifying offline RL evaluation. The reward signal is dense---engagement events occur on a daily timescale---so credit assignment is tractable. And hard safety constraints (TCPA cadence limits, time-of-day windows, consent state, escalation triggers) are easy to specify and enforce at the runtime level, exactly as the cadence-and-consent enforcement function of Section~\ref{sec:harness} requires.

Two design points generalize beyond this domain. Credit assignment across asynchronous multi-modal episodes (an ignored call Monday, an answered text Tuesday, a kept visit Wednesday) is a non-trivial problem the reward structure must address explicitly; standard RL frameworks do not solve it out of the box. And runtime-enforced safety under a non-deterministic policy---particularly when implemented through tool-calling LLM agents---requires schema validation, cadence checking, and consent verification that must execute before each action reaches the patient. The post-discharge illustration is not a complete implementation; it is a design that exhibits the architectural pattern the abstract harness specification leaves open.

\section{Empirical anchoring}
\label{sec:empirical}

\subsection{What the framework predicts}

The framework developed in Sections~\ref{sec:substrate}--\ref{sec:harness} makes several testable predictions.

\textbf{Capability-weighted offline RL outperforms uniform-weighted offline RL when training data is generated by clinicians of heterogeneous capability.} This is the load-bearing prediction of the two-loop architecture (Section~\ref{sec:twoloop}): an agent that uses $\kappa$ from the outer-loop preference learning to weight transitions in the inner-loop RL training will preferentially imitate high-capability clinicians, while an agent that ignores $\kappa$ will imitate the population mean. This prediction is tested in simulation in Section~\ref{sec:simulation} (Study A) and is confirmed: the capability-weighted variant outperforms the uniform-weighted variant under matched data, matched algorithm, and matched reward structure, with the largest gap (15-percentage-point improvement on T2D TTC) appearing in the configuration that combines capability weighting with terminal-event reward.

\textbf{Execution-intensity-aware policies generalize across deployments better than $\varepsilon$-naive policies.} A policy trained without explicit treatment of $\varepsilon$ will exhibit performance variance across deployments with different $\varepsilon$ profiles; a policy trained with the constrained MDP formulation should exhibit smaller variance because it is parameterized by $\varepsilon$. This prediction is tested in simulation in Section~\ref{sec:simulation} (Study B) and is confirmed: policies trained without $\varepsilon$ in the state cannot take advantage of higher deployment $\varepsilon$, while $\varepsilon$-aware policies adapt their recommendations to deployment regime. The deployment-side version---testing across real multi-site deployments where $\varepsilon$ varies by site characteristic---remains as a deployment-validation question.

\textbf{The harness routes a calibrated fraction of decisions to clinician review.} A well-calibrated harness should route to autonomous execution the actions where uncertainty is low and risk is below threshold, and to clinician review the actions where either condition fails. The fraction in each category is itself a measurable quantity: in deployments with mature operational infrastructure, the fraction of operational actions executing autonomously should be high (we estimate $\geq 70\%$ for routine scheduling and outreach); in clinical actions, the fraction routed for autonomous execution should be lower and grow over time as the agent's track record accumulates. This prediction is not tested in the present simulation because it requires modeling clinician interaction with the harness; we identify it as a target for prospective deployment evaluation.

\textbf{Stall-aware policies compress TTC more than stall-naive policies.} The framework's prescription that stall states (Section~\ref{sec:stalls}) trigger autonomous operational intensification predicts that policies which detect and respond to stalls will compress TTC more than policies which treat all uncontrolled states identically. The effect should be most pronounced in patient subpopulations where therapeutic inertia is most damaging---patients with multiple comorbidities, low health literacy, or under clinicians whose inferred capability $\kappa_i$ is below the population mean. This prediction is testable in deployment data where stall states can be identified retrospectively from trajectory shapes and stall-response policies can be A/B-tested prospectively.

\subsection{The empirical evidence so far}

Independent actuarial modeling, prepared by Havarti Risk and derived from review of a deployed chronic care platform's performance, indicates that compression of the uncontrolled interval (the value driver corresponding to TTC compression in the framework above) accounts for approximately 60\% of total platform value across both payer-mediated and provider-embedded deployment contexts. This is consistent with the framework's prediction that the dominant economic mechanism in chronic care platform performance is velocity compression rather than other commonly proposed value drivers (screening, quality measure improvement, capacity extension). The actuarial result does not by itself validate the RL formulation; it does establish that the reward structure proposed here is calibrated to the economic outcome that risk-bearing organizations actually capture.

The action rate assumptions in the modeling---25\% in payer-mediated contexts and 50\% in provider-embedded contexts---reflect realistic operating points for $\varepsilon$ in current deployments and provide bounds on the constraint structure of the constrained MDP for those contexts.

\subsection{Simulation results}
\label{sec:simulation}

We tested the framework's two strongest predictions on synthetic chronic care state machines: (a)~capability-weighted offline RL (the inner loop of the two-loop architecture in Section~\ref{sec:twoloop}) outperforms uniform-weighted offline RL when training data is generated by clinicians of heterogeneous capability; (b)~$\varepsilon$-aware policies generalize across deployment regimes while $\varepsilon$-naive policies do not. We summarize the design and findings here; full implementation details are in Appendix~\ref{app:sim}.

\paragraph{Design.}
Patient state machines are continuous-time biomarker processes (SBP for HTN, HbA1c for T2D) with weekly measurement, calibrated qualitatively to published treatment-response curves~\citep{whelton2018}. Each simulated population includes 2,000 patients with heterogeneous baseline severity, response coefficients, and adherence. The action space is six discrete actions (three medication intensification levels $\times$ two operational actions: do-nothing or schedule outreach). Behavior data is generated by a stochastic policy reflecting the well-documented therapeutic inertia in real clinical practice~\citep{phillips2001, okonofua2006, khunti2018}, with three clinician archetypes representing realistic heterogeneity: low-escalation clinicians (50\% of patients) with significant therapeutic inertia and rare operational action use; high-escalation clinicians (30\%) closer to guideline-directed escalation but still with rare operational action use; and operationally-augmented clinicians (20\%) representing emerging AI-augmented care delivery---high-escalation paired with active operational outreach (voice agent assistance, structured follow-through). Operational actions raise patient adherence by 0.30, within the 0.15--0.35 range supported by the active care management literature~\citep{kvedar2014}.

The agent is a tabular offline Q-learning algorithm. We hold the algorithm constant across all comparisons; only the weighting scheme and reward function differ. Each result is averaged over five seeds.

\paragraph{Study A: Capability weighting (P1 coupling).}
We compare four configurations:
\begin{itemize}
\setlength{\itemsep}{0.2em}
\item \textbf{Uniform-weighted, tiered reward}: standard offline Q-learning with the tiered TTG/TTO/TTC reward of Equation~(\ref{eq:reward}). All transitions weighted equally ($\beta = 0$).
\item \textbf{Capability-weighted, tiered reward}: same reward, but transitions weighted by $\exp(\beta \kappa_i)$ per Equation~(\ref{eq:weighting}) with $\beta = 2.5$, where $\kappa_i$ is the capability score inferred from each clinician's patient outcomes.
\item \textbf{Capability-weighted, terminal reward}: capability weighting with a sparse end-of-trajectory reward (the chronic-care analog of mortality-proxy rewards in much of the existing healthcare RL literature).
\item \textbf{Behavior policy baseline}: the heterogeneous clinician policy that generated the offline data.
\end{itemize}

\begin{figure}[t]
\centering
\includegraphics[width=0.95\textwidth]{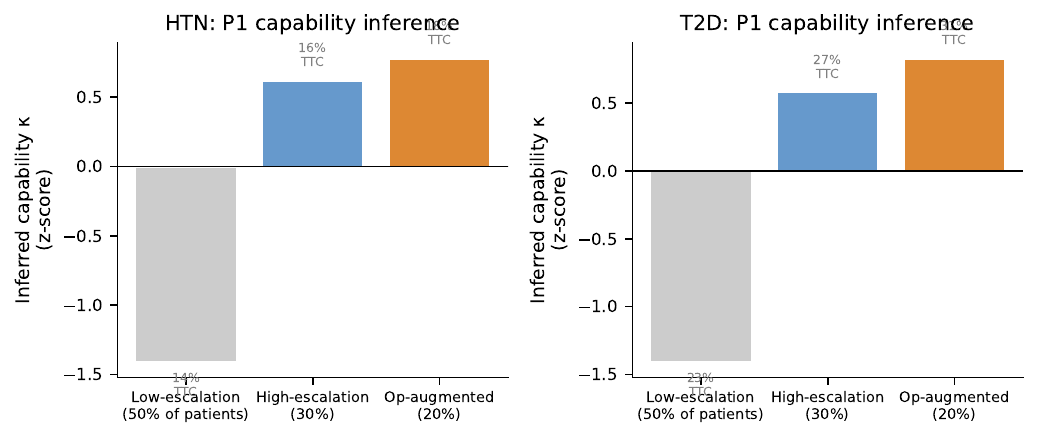}
\caption{Capability inference (outer loop). The framework correctly orders the three clinician archetypes from outcome data alone: operationally-augmented clinicians have the highest inferred capability ($\kappa = +0.78$ for HTN, $+0.83$ for T2D); low-escalation clinicians have the lowest ($\kappa = -1.41$ for both). The inner loop weights transitions by $\exp(\beta \kappa_i)$, preferentially imitating the high-capability cluster.}
\label{fig:capability_inference}
\end{figure}

The capability inference (outer loop) correctly identifies the three archetypes from outcome data alone (Figure~\ref{fig:capability_inference}): operationally-augmented clinicians achieve the highest capability score, high-escalation clinicians a moderate score, and low-escalation clinicians a strongly negative score. This matches the ground-truth ordering by which the simulation was constructed.

\begin{figure}[t]
\centering
\includegraphics[width=\textwidth]{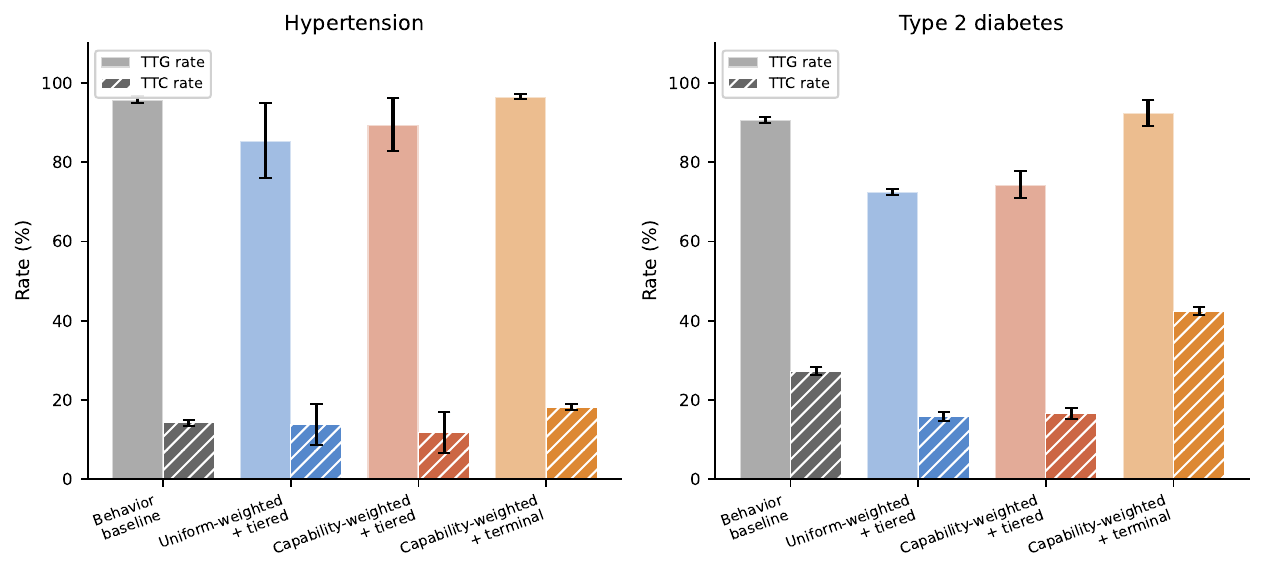}
\caption{Study A: Capability-weighted offline RL outperforms uniform-weighted offline RL and the behavior policy. The capability-weighted variant with terminal-event reward (rightmost) is the strongest configuration, beating the behavior baseline by $\sim$30\% relative on HTN TTC and $\sim$56\% on T2D TTC. The uniform-weighted variant (second from left) underperforms the behavior policy on T2D, confirming that imitating the population mean of clinician behavior produces a worse policy than the average clinician. Error bars are $\pm 1$ standard deviation across five seeds.}
\label{fig:capability}
\end{figure}

The inner loop's policy quality follows the expected pattern (Figure~\ref{fig:capability}, Table~\ref{tab:study_a}). For HTN, capability-weighted RL with terminal reward achieves $97 \pm 1\%$ TTG and $18 \pm 1\%$ TTC, exceeding the behavior policy ($96 \pm 1\%$ TTG, $14 \pm 1\%$ TTC) and outperforming uniform-weighted RL with the same reward family. For T2D, the gap is more substantial: capability-weighted RL with terminal reward achieves $42 \pm 1\%$ TTC versus $27 \pm 1\%$ for the behavior policy, a 15-percentage-point absolute improvement and a 56\% relative improvement.

\begin{table}[h]
\centering
\caption{Study A summary. TTG and TTC rates and mean biomarker reduction (mean $\pm$ std across five seeds, $n = 1{,}000$ evaluation patients per condition). Capability-weighted offline RL with terminal-event reward is the strongest configuration on both conditions and is the only configuration that exceeds behavior policy performance on T2D. Uniform-weighted offline RL underperforms the behavior policy on T2D, reflecting that imitation of the population mean of clinician behavior is a worse target than the average clinician.}
\label{tab:study_a}
\small
\begin{tabular}{lccc}
\toprule
\textbf{Configuration} & \textbf{TTG \%} & \textbf{TTC \%} & \textbf{Mean reduction} \\
\midrule
\multicolumn{4}{l}{\emph{Hypertension (mean SBP reduction in mmHg)}} \\
Behavior baseline & $96 \pm 1$ & $14 \pm 1$ & $13.8 \pm 0.3$ \\
Uniform-weighted, tiered reward & $85 \pm 10$ & $14 \pm 5$ & $11.5 \pm 2.7$ \\
Capability-weighted, tiered reward & $89 \pm 7$ & $12 \pm 5$ & $13.0 \pm 1.4$ \\
\textbf{Capability-weighted, terminal reward} & $\mathbf{97 \pm 1}$ & $\mathbf{18 \pm 1}$ & $\mathbf{14.6 \pm 0.8}$ \\
\midrule
\multicolumn{4}{l}{\emph{Type 2 diabetes (mean HbA1c reduction in \%)}} \\
Behavior baseline & $91 \pm 1$ & $27 \pm 1$ & $1.02 \pm 0.03$ \\
Uniform-weighted, tiered reward & $72 \pm 1$ & $16 \pm 1$ & $0.65 \pm 0.01$ \\
Capability-weighted, tiered reward & $74 \pm 3$ & $17 \pm 1$ & $0.69 \pm 0.07$ \\
\textbf{Capability-weighted, terminal reward} & $\mathbf{92 \pm 3}$ & $\mathbf{42 \pm 1}$ & $\mathbf{1.21 \pm 0.08}$ \\
\bottomrule
\end{tabular}
\end{table}

Two findings deserve emphasis. First, \emph{capability weighting is the load-bearing innovation}: under both reward formulations, capability-weighted variants outperform their uniform-weighted counterparts. The uniform-weighted offline Q-learning agent on T2D actually underperforms the behavior policy ($16\%$ TTC versus $27\%$), reflecting that imitation of the population mean of clinician behavior is a strictly worse target than the average clinician. The capability-weighted variant fixes this failure mode by preferentially imitating high-capability clinicians, which is exactly the mechanism the framework's two-loop architecture provides.

Second, \emph{reward formulation matters but is secondary to capability weighting}. The terminal-event reward produces the strongest policies in this setup, despite its lower signal density (Proposition~\ref{prop:density}), because end-of-trajectory outcome is a clean, unambiguous training signal that does not over-credit intermediate progress. The tiered reward provides denser credit assignment but appears to over-reward the agent for partial responses (TTG hits at 15~mmHg reduction) that do not necessarily lead to full control, leaving the agent with less incentive to push for the higher-reward TTC milestone. This is a tractable problem for richer state representations and reward shaping methods we leave to subsequent work, but the clean finding is that for the simple tabular setup tested here, terminal-event rewards under capability weighting produce the best policies. The tiered reward's value as the framework's reward structure derives from its alignment with the ACCESS payment structure (Section~\ref{sec:reward}) and from the credit-assignment density argument of Proposition~\ref{prop:density}; richer experimental settings are likely needed to show its full advantage.

\paragraph{Study B: Execution-intensity generalization.}
We tested whether $\varepsilon$-aware policies generalize across deployment regimes. Two policy variants were trained: an $\varepsilon$-naive policy on data from $\varepsilon = 0.5$ only, and an $\varepsilon$-aware policy that includes $\varepsilon$ as a state variable, trained on data drawn from $\varepsilon \in \{0.25, 0.5, 0.75\}$ with matched total volume. Both were evaluated at four deployment values $\varepsilon \in \{0.25, 0.5, 0.75, 0.9\}$.

\begin{figure}[t]
\centering
\includegraphics[width=\textwidth]{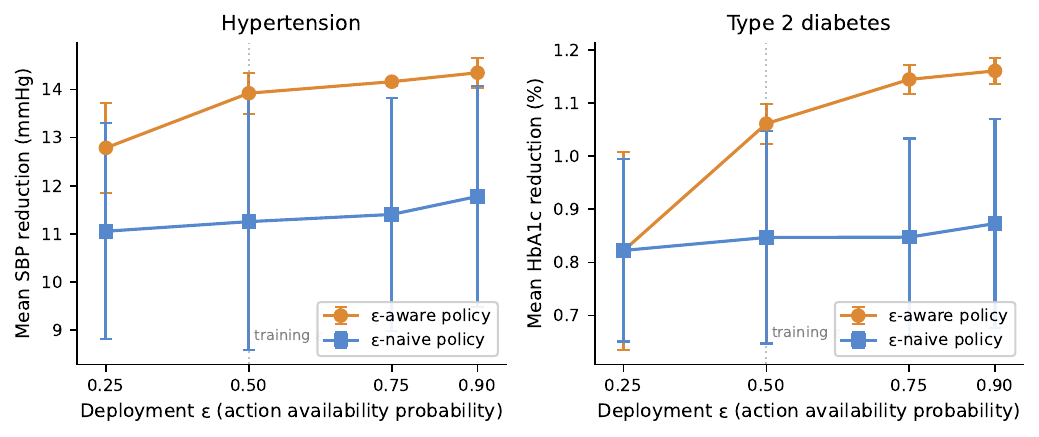}
\caption{Study B: Execution-intensity generalization. The $\varepsilon$-aware policy adapts to higher deployment $\varepsilon$ by recommending more aggressively. The $\varepsilon$-naive policy stays approximately flat across deployment regimes; it cannot take advantage of higher action availability because it was trained on a single $\varepsilon$ and specialized to that operating point. Error bars are $\pm 1$ standard deviation across three seeds.}
\label{fig:study_b}
\end{figure}

The $\varepsilon$-aware policy outperforms the $\varepsilon$-naive policy at every deployment $\varepsilon$ (Figure~\ref{fig:study_b}). At the matched $\varepsilon = 0.5$ deployment, mean SBP reduction is $13.9 \pm 0.4$ mmHg for the aware policy versus $11.3 \pm 2.7$ for the naive policy; for T2D, $1.06 \pm 0.04\%$ versus $0.85 \pm 0.10\%$. More substantively, the $\varepsilon$-aware policy adapts to higher deployment $\varepsilon$ while the $\varepsilon$-naive policy does not. As deployment $\varepsilon$ rises from 0.5 to 0.9, the aware policy's HTN performance climbs from 13.9 to 14.3 mmHg, and its T2D performance climbs from $1.06\%$ to $1.16\%$. The naive policy stays approximately flat: $11.3$ to $11.8$ mmHg for HTN, $0.85\%$ to $0.87\%$ for T2D. This is the deployment-simulation gap of Section~\ref{sec:eps} made empirically concrete: a policy that bakes in fixed assumptions about action availability does not generalize, even when the deployment is favorably shifted.

\paragraph{Limitations of the simulation.}
The simulation is structurally faithful but quantitatively stylized. Patient state machines are calibrated qualitatively to published response curves rather than fit to deployment data. The tabular state representation is coarse, which caps the achievable absolute policy quality below what richer representations would achieve. The behavior policy is a constructed three-cluster mixture rather than a calibrated representation of any specific clinical practice. The capability inference exploits the fact that the simulation has only three archetypes with hundreds of patients each; in real deployment, capability is estimated per-clinician across smaller per-clinician sample sizes, with correspondingly higher uncertainty. The simulation tests the structural claims of the framework---capability weighting beats uniform weighting, $\varepsilon$-awareness produces generalization---but does not validate quantitative deployment claims. We treat the simulation as confirmation that the structural arguments hold under controlled conditions; deployment validation is the natural next step.

\subsection{What requires deployment}

Several claims cannot be settled by simulation alone and require prospective deployment evaluation. The harness's routing calibration in real clinical workflow, the actual rate at which $\varepsilon$ rises as the agent accumulates clinician interaction data, the speed at which capability scores converge given realistic per-clinician sample sizes, and the performance of the framework when integrated with concrete engagement-harness instantiations of the kind described in Section~\ref{sec:harness} all require deployed observation. We treat these as the natural successor research program rather than as claims of the present paper.

\section{Discussion}
\label{sec:discussion}

\subsection{Relationship to companion work}

The framework's central technical claim is its tight coupling to the preference-learning machinery of~\citet{altitudep1}. That paper is not cited here as background; it is the source of two quantities ($\varepsilon$ and $\kappa$) that enter our framework as load-bearing structural elements. The Bradley-Terry preference model in \citet{altitudep1} infers per-clinician capability $\kappa_i$ from outcome data accumulated across the clinician's panel; the formal decomposition $\varepsilon = \varepsilon_{\mathrm{cap}} \cdot \varepsilon_{\mathrm{will}}$ identifies execution intensity as a state variable rather than an exogenous nuisance. We use both: $\varepsilon$ in the action-availability constraint, $\kappa$ in the offline RL training objective. The simulation in Section~\ref{sec:simulation} shows that the coupling is empirically load-bearing: a chronic care RL agent that uses these quantities outperforms one that ignores them.

This bidirectional coupling between preference learning and RL is, to our knowledge, the first such formulation specifically for chronic care. It addresses an open problem in healthcare RL: the unidentified behavior policy that has bedeviled offline RL evaluation since the early sepsis work~\citep{komorowski2018, gottesman2019}. By modeling clinician-level variation explicitly through $\kappa$, the framework neither requires nor pretends to a uniform behavior policy assumption.

\subsection{Open problems}

Several open problems are worth naming explicitly.

\textbf{Capability inference under realistic per-clinician sample sizes.} The capability score $\kappa_i$ is a posterior estimate from outcome data; in our simulation each archetype has hundreds of patients and the inference is precise. In real deployment, individual clinicians have smaller panels and the posterior over $\kappa_i$ has higher variance. The temperature $\beta$ in the weighting term must be calibrated to this uncertainty; an overconfident $\beta$ on noisy capability estimates will produce a policy that imitates clinicians who happen to have favorable outcomes by chance. Posterior-aware weighting (e.g., scaling $\beta$ by the inverse posterior variance per clinician) is a natural extension we leave to subsequent work.

\textbf{Identifiability of $\varepsilon_{\mathrm{cap}}$ and $\varepsilon_{\mathrm{will}}$.} The constraint structure of Section~\ref{sec:eps} requires the agent to estimate execution intensity as a state variable. The decomposition into capacity and willingness components is informative for action selection but is not always point-identified from observed data, as discussed in \citet{altitudep1}. Partial identification approaches and natural-experiment designs that move one component while holding the other constant are open methodological questions.

\textbf{Two-loop convergence guarantees.} The two-loop architecture of Section~\ref{sec:twoloop} couples preference learning to RL, but the joint convergence properties of the coupled loops are not formally characterized. The outer loop's posterior over $(\varepsilon, \kappa)$ updates as new override data accumulates; the inner loop's policy updates as retraining completes; deployments of the inner loop's policy generate new override data that updates the outer loop's posterior. Conditions under which this coupled system converges to a fixed point---and rates of convergence---are open theoretical questions.

\textbf{Off-policy evaluation under typed actions.} The two-layer action taxonomy we adopt (Section~\ref{sec:substrate}) refines the action space in a way that should improve off-policy evaluation by reducing within-action heterogeneity. The formal evaluation properties of typed-action OPE under standard offline RL methods is an open question.

\textbf{Harness routing function learning.} We have specified the harness's routing function abstractly but have not formalized its learning. The routing function should itself be subject to update as the agent's track record accumulates, but the conditions under which the harness can safely move decisions from category 3 to 1 require formal treatment.

\subsection{Limitations}

The framework's empirical validation here consists of the actuarial modeling in Section~\ref{sec:empirical} and the simulation in Section~\ref{sec:simulation}. The simulation uses synthetic state machines calibrated qualitatively to published response curves, a coarse tabular state representation, and a constructed three-cluster behavior policy; capability inference exploits the simulation's three-archetype structure with hundreds of patients per archetype, whereas real deployment estimates capability per clinician on smaller panels with correspondingly higher uncertainty. These choices are appropriate for testing the structural argument but do not validate quantitative deployment claims. Retrospective evaluation on real deployment data and prospective deployment study are the natural successor experiments and are out of scope here.

A second deployment dependency is the patient state encoder $\phi(s)$ that maps the conceptual clinical state to the feature vector the learning algorithm operates on. A production $\phi(s)$ is high-dimensional and structured (raw biomarker values, derived control states, comorbidity flags, adherence signals, symptom features, social and structural determinants, observation density, operational metadata, provider attribution); building a high-quality $\phi(s)$ is the central ML engineering effort in production deployment and bounds deployed quality. The framework's structural arguments hold at the level of the formal state space and constrain $\phi(s)$ (the layers must each be representable; $\kappa$ must be identifiable; $\varepsilon$ must be estimable) but do not solve the standard offline ML problems of missing data, dimensionality, and feature selection.

We have not engaged with the algorithmic choice of offline RL method. Conservative methods~\citep{kumar2020}, batch-constrained methods~\citep{fujimoto2019}, implicit Q-learning~\citep{kostrikov2022}, and hindsight-based methods~\citep{andrychowicz2017} all have arguments for different aspects of the problem; recent offline-RL work on type~1 diabetes blood-glucose management~\citep{emerson2023} is encouraging for the chronic care setting more broadly. The relative differences between weighted and unweighted variants observed under tabular Q-learning should hold under stronger algorithms, though function approximation introduces failure modes the tabular setting does not---particularly poor Q estimates for underrepresented state-action pairs---and data-completeness requirements for neural offline RL on chronic care data are an open empirical question. A related specific finding: the tiered reward did not dominate the terminal-event reward as cleanly as Proposition~\ref{prop:density} predicts, which we attribute to the tabular representation over-crediting partial responses; the tiered reward's value derives primarily from its ACCESS alignment, and demonstrating its full advantage likely requires richer experimental settings.

The framework is specific to chronic care under value-based payment structures; extensions to acute care, surgical decision-making, or other healthcare RL settings would require substantial reformulation.

\section{Conclusion}

Chronic disease management is a structurally favorable domain for reinforcement learning, but only if the reward signal, action space, constraint structure, and offline-data weighting scheme are formalized in a way that exploits the domain's properties rather than importing structures from acute-care RL that fit poorly. The TTG/TTO/TTC reward hierarchy provides tiered, contractually-anchored credit assignment calibrated to the ACCESS payment structure; the two-layer action taxonomy separates clinical from operational decisions in a way that aligns with the causal structure of outcomes; the execution-intensity-constrained MDP makes deployable performance, not simulation performance, the optimization target; and the supervisory harness bounds agent autonomy while permitting autonomous execution of routine operational decisions.

The framework's distinctive feature is its tight coupling to clinician preference learning. Execution intensity $\varepsilon$ and clinician capability $\kappa$, derived in \citet{altitudep1}, enter our formulation as load-bearing structural elements. In our simulation, capability-weighted offline RL outperforms both uniform-weighted offline RL and the behavior policy baseline, while uniform-weighted offline RL---the standard formulation in the existing literature---underperforms even the heterogeneous behavior policy on T2D. The empirical signature of a real coupling: imitation of the population mean is a worse target than the average clinician, and the framework's preference-weighted alternative fixes this failure mode. Deployment-side validation is the next phase of work.

\bibliographystyle{plainnat}
\bibliography{ms}

\appendix

\section{Simulation implementation details}
\label{app:sim}

\subsection{Patient state machines}

For hypertension, each patient $i$ is parameterized by a setpoint $\mu_i \sim \mathcal{N}(160, 12)$ clipped to $[135, 195]$ mmHg, two response coefficients $r^{(1)}_i \sim \mathcal{N}(10, 2.5)$ and $r^{(2)}_i \sim \mathcal{N}(20, 4)$ clipped at floor values, and an adherence rate $\alpha_i \sim \mathrm{Beta}(7, 3)$. At each weekly timestep $t$, given current medication level $m \in \{0, 1, 2\}$ and weeks-on-current-medication $w$, let $X_t \sim \mathrm{Bernoulli}(\alpha_{\mathrm{eff},t})$ indicate whether the patient takes their medication this week, where $\alpha_{\mathrm{eff},t} = \min(\alpha_i + 0.30 \cdot a_{\mathrm{op},t}, 0.98)$ and $a_{\mathrm{op},t} \in \{0, 1\}$ is the operational action. SBP is generated as
\begin{equation}
\mathrm{SBP}_i(t) = \mu_i - r^{(m)}_i \cdot (1 - e^{-w/4}) \cdot X_t + \epsilon_t, \qquad \epsilon_t \sim \mathcal{N}(0, 4)
\end{equation}
where the operational-action adherence boost of 0.30 reflects AI-augmented operational intensity (voice agent assistance, structured outreach), within the 0.15--0.35 range supported by active care management adherence literature~\citep{kvedar2014}.

For type 2 diabetes, the analogous parameterization is $\mu_i \sim \mathcal{N}(8.8, 1.0)$ clipped to $[7.2, 12.5]\%$, with response coefficients $r^{(1)}_i \sim \mathcal{N}(0.9, 0.25)$ and $r^{(2)}_i \sim \mathcal{N}(1.8, 0.4)$. The medication ramp-in time constant is 8 weeks, reflecting the slower response of HbA1c to therapy. Measurement noise is $\epsilon_t \sim \mathcal{N}(0, 0.15)$ in HbA1c units.

\subsection{Milestone definitions}

For each patient, baseline is the SBP or HbA1c value at $t=0$ before any intervention. Milestones are first-passage events:

\begin{itemize}
\item HTN reduction-from-baseline thresholds: $\mathrm{TTG}$ at $\geq 15$ mmHg reduction (i.e., first time $\mathrm{SBP}(t) \leq \mathrm{baseline} - 15$); $\mathrm{TTO}$ at $\geq 25$ mmHg reduction; $\mathrm{TTC}$ requires SBP $< 130$ mmHg sustained for the previous 4 consecutive weeks (4-week confirmation window).
\item T2D reduction-from-baseline thresholds: $\mathrm{TTG}$ at $\geq 1.0\%$ HbA1c reduction; $\mathrm{TTO}$ at $\geq 1.5\%$ HbA1c reduction; $\mathrm{TTC}$ requires HbA1c $< 7.0\%$ sustained for the previous 4 consecutive weeks.
\end{itemize}

\subsection{Behavior policy}

The behavior policy generating the offline data is a three-cluster mixture reflecting the heterogeneity of real clinical practice. Each patient is assigned to a clinician archetype:

\begin{itemize}
\setlength{\itemsep}{0.2em}
\item \textbf{Low-escalation (50\% of patients)}: significant therapeutic inertia~\citep{phillips2001, okonofua2006, khunti2018}. Initial escalation probability when uncontrolled is $\min(0.10 + 0.02 w, 0.50)$; escalation to second-line is $\min(0.05 + 0.015(w-8), 0.25)$ after $w \geq 8$ weeks on first-line. Operational action use is 5\%.
\item \textbf{High-escalation (30\%)}: closer to guideline-directed escalation. Initial escalation $\min(0.20 + 0.04 w, 0.70)$; second-line $\min(0.15 + 0.025(w-6), 0.45)$. Operational action use is 5\%.
\item \textbf{Operationally-augmented (20\%)}: high-escalation paired with active operational outreach. Same escalation rates as high-escalation cluster, but operational action use is 45\% during uncontrolled visits and 10\% during controlled visits. Represents emerging AI-augmented care delivery (voice agent assistance, structured outreach).
\end{itemize}

The T2D policy follows the same structure with adjusted timing constants reflecting slower glycemic response (12-week minimum on first-line before second-line consideration in high cluster; 16-week minimum in low cluster). Operational actions raise patient adherence by 0.30 per week, with adherence capped at 0.98.

\subsection{State discretization}

For the offline Q-learning agent, state is discretized into a tuple of integer buckets:
\begin{itemize}
\item For HTN: SBP into 10 buckets (110--200 mmHg in 10 mmHg increments), current medication (3 levels), weeks-on-current bucket (0--3, 4--7, 8+), and reduction-from-baseline bucket (no response, partial, TTG-hit, substantial). The reduction bucket is critical: without it, the agent cannot distinguish a working response from a non-response and learns degenerate policies.
\item For T2D: HbA1c into 12 buckets (6.0--12.0\% in 0.5\% increments), with analogous medication, weeks-on, and reduction-from-baseline buckets.
\end{itemize}
For the $\varepsilon$-aware variant, an additional 4-bucket discretization of $\varepsilon$ is added to the state.

\subsection{Algorithm}

We use a tabular offline Q-learning agent with the standard update
\begin{equation}
Q_{t+1}(s, a) = Q_t(s, a) + \eta \big[ r + \gamma \max_{a'} Q_t(s', a') - Q_t(s, a) \big]
\end{equation}
where $\eta = 0.05$ is the learning rate and $\gamma = 0.97$ the discount factor. The configurations differ in how transitions are sampled to form training batches:
\begin{itemize}
\setlength{\itemsep}{0.2em}
\item \textbf{Uniform-weighted}: transitions are sampled uniformly from the offline dataset (equivalent to $w_i = 1$ for all $i$ in Equation~(\ref{eq:weighting})).
\item \textbf{Capability-weighted}: transitions are sampled with probability proportional to $w_i = \exp(\beta \kappa_{c(i)})$ where $c(i)$ is the clinician archetype for patient $i$ and $\kappa_c$ is the inferred capability score for archetype $c$. We use $\beta = 2.5$ in Study~A. Sampling proportionally to $w_i$ is the importance-sampling implementation of the weighted objective in Equation~(\ref{eq:objective}); the per-transition Q update remains the standard form above without an explicit $w_i$ factor, so weighting is not double-applied.
\end{itemize}

The capability scores $\kappa_c$ are computed from patient outcomes: each patient's outcome score is mean biomarker reduction plus a 5-unit bonus for TTC achievement (scaled to compatible units across HTN and T2D). Per-archetype mean scores are then z-normalized across the three archetypes. In all five seeds and both conditions, the inferred ordering recovers the ground-truth ordering: operationally-augmented \(>\) high-escalation \(\gg\) low-escalation, with $\kappa$ values approximately $\{+0.8, +0.6, -1.4\}$.

We train for 600 iterations with batch size 512.

\subsection{Reward functions}

Two reward functions are used in Study A. Both share the action cost decomposition
\begin{equation}
\rho(a_t, a_{t-1}) = 0.01 \cdot \mathbb{1}[a_{\mathrm{med},t} \neq a_{\mathrm{med},t-1}] + 0.005 \cdot \mathbb{1}[a_{\mathrm{op},t} = 1]
\end{equation}
where the first term is a medication-change cost and the second is an operational-action cost; both costs apply additively when both conditions hold. Indicator functions on milestone events fire only at the unique time of first passage:
\begin{align}
R_{\mathrm{tiered}}(t) &= 1.0 \cdot \mathbb{1}[t = T_0 + \mathrm{TTG}] + 1.5 \cdot \mathbb{1}[t = T_0 + \mathrm{TTO}] \nonumber \\
&\quad + 2.5 \cdot \mathbb{1}[t = T_0 + \mathrm{TTC}] - \rho(a_t, a_{t-1}) \\
R_{\mathrm{terminal}}(t) &= \mathbb{1}[t = T_{\mathrm{end}}] \cdot \big( 2.5 \cdot \mathbb{1}[\mathrm{controlled\ at\ } T_{\mathrm{end}}] \nonumber \\
&\quad - 2.5 \cdot \mathbb{1}[\mathrm{poor\ outcome\ at\ } T_{\mathrm{end}}] \big) - \rho(a_t, a_{t-1})
\end{align}
The simulation uses tiered weights $(w_G, w_O, w_C) = (1.0, 1.5, 2.5)$ rather than the $w_G \approx w_C$ calibration motivated in Section~\ref{sec:reward}. We chose strict ordering $w_G < w_O < w_C$ for the simulation because in a synthetic environment without contractual payment alignment, the credit-assignment-density argument is what matters for testing, and strict ordering provides cleaner ordering of intermediate vs full-control achievement. The contractual-alignment calibration $w_G \approx w_C$ from ACCESS is more relevant for deployment than for this simulation. The terminal-event reward is the chronic-care analog of mortality-proxy rewards used in much of the existing healthcare RL literature, included as a comparison anchor.

\end{document}